\documentclass[10pt,twocolumn,letterpaper]{article}

\usepackage[pagenumbers]{cvpr} 

\usepackage{graphicx}
\usepackage{amsmath}
\usepackage{amssymb}
\usepackage{booktabs}
\usepackage{indentfirst}
\usepackage{color,soul}
\usepackage{color,colortbl}
\usepackage{multirow}
\usepackage{algorithm}  
\usepackage{algorithmicx}  
\usepackage{algpseudocode}  
\usepackage{amsmath} 
\usepackage{paralist}
\usepackage[table]{xcolor}
\floatname{algorithm}{{Algorithm}}

\definecolor{lightgray}{gray}{0.9}
\definecolor{LightCyan}{rgb}{0.88,1,1}
\newcommand{\mysize}{1.6in} 
\usepackage[labelformat=simple]{subcaption}

\usepackage[pagebackref=true,breaklinks=true,letterpaper=true,colorlinks,citecolor=blue,linkcolor=blue,bookmarks=false]{hyperref}

\usepackage[capitalize]{cleveref}
\crefname{section}{Sec.}{Secs.}
\Crefname{section}{Section}{Sections}
\Crefname{table}{Table}{Tables}
\crefname{table}{Tab.}{Tabs.}

\def\cvprPaperID{0000} 
\def\confName{CVPR}
\def\confYear{2023}

\begin{document}

\title{Exploiting Completeness and Uncertainty of Pseudo Labels\\ for Weakly Supervised Video Anomaly Detection}
	
\author{Chen Zhang$^{1,2}$ \quad Guorong Li$^{3,*}$ \quad Yuankai Qi$^{4}$ \quad Shuhui Wang$^{5,6}$ \\ Laiyun Qing$^{3}$ \quad Qingming Huang$^{3,5,6}$ \quad Ming-Hsuan Yang$^{7}$ \\
	$^{1}$State Key Laboratory of Information Security, Institute of Information Engineering, CAS \\
	$^{2}$School of Cyber Security, University of Chinese Academy of Sciences \\
	$^{3}$School of Computer Science and Technology, University of Chinese Academy of Sciences \\
	$^{4}$Australian Institute for Machine Learning, The University of Adelaide \\
	$^{5}$Key Laboratory of Intelligent Information Processing, Institute of Computing Technology, CAS \\
	$^{6}$Peng Cheng Laboratory, $^{7}$University of California, Merced \\
	{\tt\small zhangchen@iie.ac.cn}, {\tt\small liguorong@ucas.ac.cn}, {\tt\small qykshr@gmail.com}, 
	\\ {\tt\small wangshuhui@ict.ac.cn}, {\tt\small \{lyqing, qmhuang\}@ucas.ac.cn}, {\tt\small mhyang@ucmerced.edu}
}
	\maketitle
	
	\begin{abstract}
Weakly supervised video anomaly detection aims to identify abnormal events in videos using only video-level labels. 
Recently, two-stage self-training methods have achieved significant improvements by self-generating pseudo labels and self-refining anomaly scores with these labels.
As the pseudo labels play a crucial role, we propose an enhancement framework by exploiting completeness and uncertainty properties for effective self-training. 
Specifically, we first design a multi-head classification module (each head serves as a classifier) with a diversity loss to maximize the distribution differences of predicted pseudo labels across heads. 
This encourages the generated pseudo labels to cover as many abnormal events as possible. 
We then devise an iterative uncertainty pseudo label refinement strategy, which improves not only the initial pseudo labels but also the updated ones obtained by the desired classifier in the second stage. 
Extensive experimental results demonstrate the proposed method performs favorably against state-of-the-art approaches on the UCF-Crime, TAD, and XD-Violence benchmark datasets.
	\end{abstract}
	
	\section{Introduction}
	\label{sec:intro}
	
	Automatically detecting abnormal events in videos has attracted increasing attention for its broad applications in intelligent surveillance systems.  
Since abnormal events are sparse in videos, recent studies are mainly developed within the weakly supervised learning framework~\cite{sultani2018real, zhang2019temporal, zhong2019graph, zhu2019motion, zaheer2020claws, feng2021mist, tian2021weakly, wu2020not, wu2021learning, lv2021localizing, li2022self, sapkota2022bayesian}, where only video-level annotations are available. However, the goal of anomaly detection is to predict frame-level anomaly scores during test.
This results in great challenges for weakly supervised video anomaly detection. 
	
	\begin{figure}[!t]
		\centering
        \includegraphics[width=\linewidth]{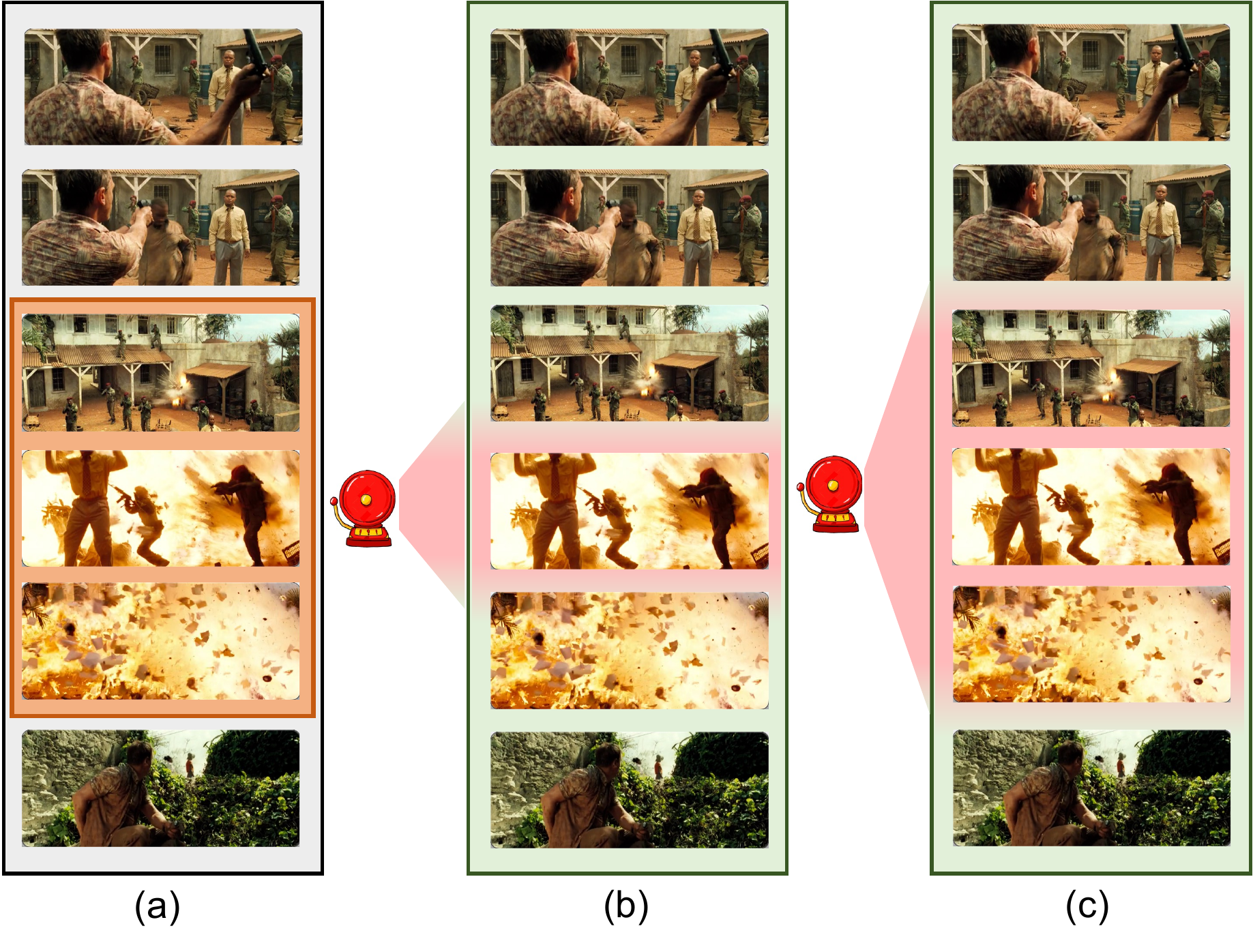}
		\caption{
        Illustration of the completeness: (a) represents a video that contains multiple abnormal clips (ground truth are in orange area). Existing methods tend to focus on the most anomalous clip as shown in (b), we propose to use the multi-head classification module together with a diversity loss to encourage pseudo labels to cover the complete abnormal events as depicted in (c). 
		}
		\label{fig:motivation}
	\end{figure}

	Existing methods broadly fall into two categories:  one-stage methods based on Multiple Instance Learning (MIL) and two-stage self-training methods.
	One-stage MIL-based methods \cite{sultani2018real, zhang2019temporal, zhu2019motion, tian2021weakly, lv2021localizing} treat each normal and abnormal video as a negative and positive bag respectively, and clips of a video are the instances of a bag. 
	Formulating anomaly detection as a regression problem (1 for abnormal and 0 for normal events), these methods adopt ranking loss to encourage the highest anomaly score in a positive bag to be higher than that in a negative bag.
	Due to the lack of clip-level annotations, the anomaly scores generated by MIL-based methods are usually less accurate. 
 	To alleviate this problem, two-stage self-training methods are proposed~\cite{feng2021mist,li2022self}. In the first stage, pseudo labels for clips are generated by MIL-based methods.
 	In the second stage, MIST~\cite{feng2021mist} utilizes these pseudo labels to refine discriminative representations. 
 	In contrast, MSL~\cite{li2022self} refines the pseudo labels via a transformer-based network. Despite progress, existing methods still suffer two limitations.
	First, the ranking loss used in the pseudo labels generation ignores the completeness of abnormal events. 
	The reason is that a positive bag may contain multiple abnormal clips as shown in Figure \ref{fig:motivation}, but MIL is designed to detect only the most likely one.
	The second limitation is that the uncertainty of generated pseudo labels is not taken into account in the second stage.
	As the pseudo labels are usually noisy, directly using them to train the final classifier may hamper its performance. 

    To address these problems, we propose to enhance pseudo labels via exploiting completeness and uncertainty properties. 
	Specifically, to encourage the complete detection of abnormal events, we propose a multi-head module to generate pseudo labels (each head serves as a classifier) and introduce a diversity loss to ensure the distribution difference of pseudo labels generated by the multiple classification heads. 
	In this way, each head tends to discover a different abnormal event, and thus the pseudo label generator covers as many abnormal events as possible.
	Then, instead of directly training a final classifier with all pseudo labels, we design  an iterative uncertainty-based training strategy. We measure the uncertainty using Monte Carlo (MC) Dropout \cite{gal2016dropout} and only clips with lower uncertainty are used to train the final classifier. At the first iteration, we use such uncertainty to refine pseudo labels obtained in the first stage, and in the remaining iterations, we use it to refine the output of the desired final classifier.

The main contributions of this paper are as follows:
\begin{compactitem}
	    \item We design a multi-head classifier scheme together with a diversity loss to encourage the pseudo labels to cover as many abnormal clips as possible.

	    \item We design an iterative uncertainty-aware self-training strategy to gradually improve the quality of pseudo labels. 

	    \item Experiments on UCF-Crime, TAD, and XD-Violence datasets demonstrate the favorable performance compared to several state-of-the-art methods.
\end{compactitem}

\section{Related Work}
\label{sec:Related}
	
\noindent	\textbf{Semi-Supervised Methods.} In the semi-supervised setting, only normal videos are required in training set. Semi-supervised video anomaly detection methods can be divided into one-class classifier-based methods \cite{xu2015learning, sabokrou2018adversarially, wu2019deep}, and reconstruction-based methods \cite{cong2011sparse, ren2015unsupervised, hasan2016learning, chong2017abnormal, park2020learning, cai2021appearance}. In one-class classifier-based methods, the model constructs a boundary that distinguishes normal events from abnormal events by learning information about normal videos. Xu \emph{et al.} \cite{xu2015learning} adopt an autoencoder to learn appearance and motion features as well as their associations, and then use multiple one-classifiers to predict anomaly scores based on these three learned feature representations. Sabokrou \emph{et al.} \cite{sabokrou2018adversarially} propose to train an end-to-end one-classification model in an adversarial manner, which can be applied to video anomaly detection. For the problem of anomaly detection in complex scenarios, Wu \emph{et al.} \cite{wu2019deep} propose to jointly optimize representation learning and one-class classification using convolutional neural networks. Reconstruction-based methods aim to minimize the reconstruction error of the training data, and takes the minimum error as the threshold for discriminating abnormality. Some \cite{cong2011sparse, ren2015unsupervised, hasan2016learning} learn the dictionary of normal videos, and video clips that cannot be represented by the dictionary are determined to be abnormal. The others \cite{hasan2016learning, chong2017abnormal} learn the rules of normal video sequences through autoencoders, and events that produce higher reconstruction errors are judged as anomalies. To prevent the model from reconstructing normal videos too well, later works \cite{park2020learning, cai2021appearance, liu2021hybrid} introduce a memory module for recording normal patterns.
	
\noindent \textbf{Weakly Supervised Methods.} Different from the semi-supervised setting, there are both normal and abnormal videos in the training set for weakly supervised methods, but frame-level annotations are not available. Most of the weakly supervised anomaly detection methods are one-stage MIL-based methods. In \cite{sultani2018real}, the first MIL-based method with ranking loss is proposed, along with a large-scale video anomaly detection dataset. Later, Zhang \emph{et al.} \cite{zhang2019temporal}  propose an inner bag loss, which is complementary to the outer bag ranking loss. To learn a motion-aware feature that can better detect anomalies, Zhu \emph{et al.} \cite{zhu2019motion} use an attention module to take temporal context into the multi-instance ranking model.  Tian \emph{et al.} \cite{tian2021weakly} develop a robust top-$k$ MIL method for weakly supervised video anomaly detection by training a temporal feature magnitude learning function. To effectively utilize the temporal context, Lv \emph{et al.} \cite{lv2021localizing} propose a high-order context encoding model. Sapkota \emph{et al.} \cite{sapkota2022bayesian} construct a submodularity diversified MIL loss in a Bayesian non-parametric way, which can satisfy anomaly detection in more realistic settings with outliers and multimodal scenarios. 

\begin{figure*}[!t]
        \centerline{\includegraphics[width=\textwidth]{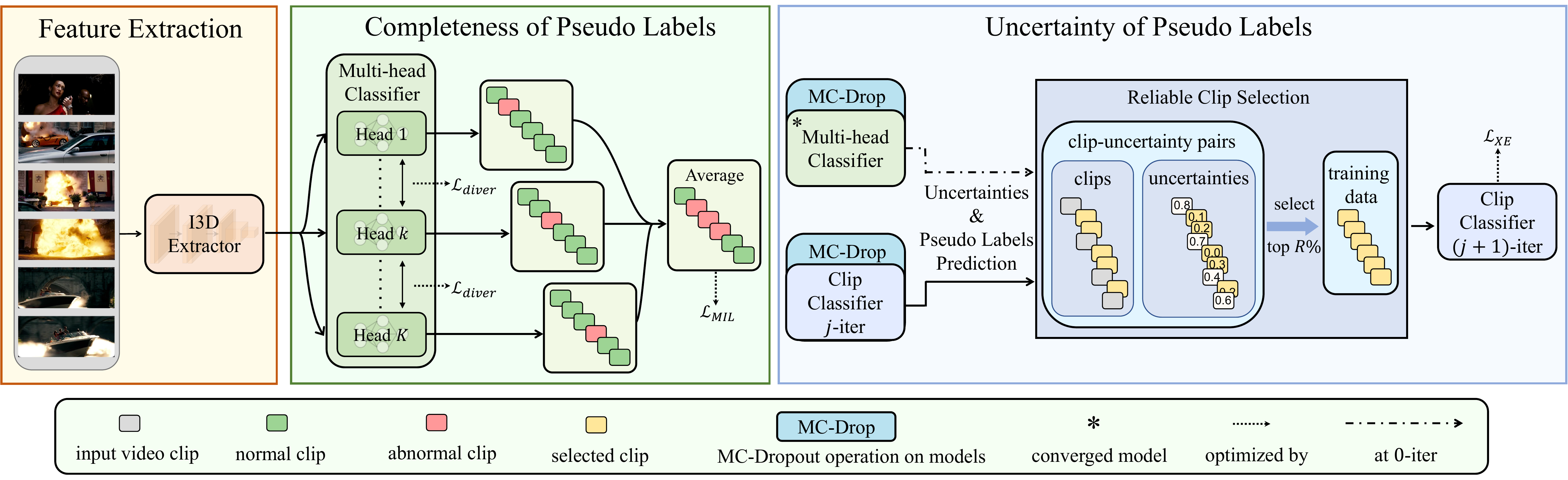}}
		\caption{Pipeline of the proposed method: (1) Completeness Enhanced Pseudo Label Generator (Sec.~\ref{sec:com}):  First, we use the pre-trained 3D CNN  to extract video features. Then the features are fed into the multi-head classifier constrained by a diversity loss to encourage the detection of complete abnormal events. Simultaneously, the MIL ranking loss is used to constrain the anomaly scores of abnormal segments to be larger than those of normal segments. (2) Iterative Uncertainty Aware Pseudo Label Refinement (Sec.~\ref{sec:uncer}):
		In the first iteration, 
		we obtain the initial clip-level pseudo labels from the multi-head classifier of the first stage and compute its uncertainty  via Monte Carlo Dropout. Then we select reliable clips based on the uncertainty to train a new clip classifier. In the remaining iterations use the new clip classifier to update pseudo labels.
		}
		\label{fig:framework}
	\end{figure*}
Recently, two-stage self-training methods are proposed to generate more accurate and fine-grained anomaly scores, which adopt a two-stage pipline, \emph{i.e.}, pseudo labels are generated first and then fed into a classification module. Feng \emph{et al.} \cite{feng2021mist} propose to use the information provided by the multi-instance pseudo label generator to fine-tune the feature encoder to generate more discriminative features specifically for the task of video anomaly detection. Li \emph{et al.} \cite{li2022self} select the sequence consisting of multiple instances with the highest sum of anomaly score as the optimization unit of ranking loss, and gradually reduced the length of the sequence by adopting a self-training strategy to refine the anomaly scores. In addition to the above two categories of methods, there are some fancy ideas for weakly supervised video anomaly detection. Zhong \emph{et al.}\cite{zhong2019graph} reformulate the weakly supervised anomaly detection problem as a supervised learning task under noisy labels, and gradually generated clean labels for the action classifier through a designed graph convolutional network.  Wu \emph{et al.} \cite{wu2020not} propose an audio-visual dataset and design a holistic and localized framework to explicitly model relations of video snippets to learn discriminative feature representations. In \cite{wu2021learning}, Wu \emph{et al.} further explore the importance of temporal relation and discriminative features for  weakly supervised anomaly detection.

\noindent \textbf{Self-Training.} Self-training is one of the mainstream techniques in semi-supervised learning \cite{lee2013pseudo, rizve2021defense, wei2021crest} and has recently shown important progress for tasks like classification \cite{xie2020self, mukherjee2020uncertainty, gavrilyuk2021motion} and domain adaptation \cite{zou2019confidence, liu2021cycle}. For self-training, the training data consists of a small amount of labeled data and a large amount of unlabeled data. The core idea is to use the model trained with labeled data to generate pseudo labels of unlabeled data, and then train the model jointly with labeled data and pseudo labels. The training process is repeated until the model converges. In weakly supervised video anomaly detection, Feng \emph{et al.} \cite{feng2021mist} propose a self-training framework in which clip-level pseudo labels generated by a multiple instance pseudo label generator are assigned to all clips of abnormal videos to refine feature encoder. Most similar to our work is the Multi-Sequence Learning method proposed by Li \emph{et al.} \cite{li2022self}, which refines anomaly scores by gradually reducing the length of selected sequences by self-training. However, the self-training mechanisms used by these methods do not consider the uncertainty of pseudo labels, leading to a gradually deviating self-training process guided by noisy pseudo labels.
In contrast, we develop an uncertainty aware self-training strategy that can reduce the effect of unreliable pseudo labels. We also consider the importance of temporal context for video understanding, in an effort to better refine anomaly scores.

	\section{Method}
The proposed pseudo label enhancement framework based on completeness and uncertainty is shown in Figure~\ref{fig:framework}. 
We first use a multi-head classifier trained with a diversity loss and MIL ranking loss to generate   initial clip-level pseudo labels. Then, we utilize an iterative uncertainty aware pseudo label refinement strategy  to gradually improve the quality of pseudo labels to train the final desired classifier. 
In the following, we first formulate the task of weakly supervised video anomaly detection and then elaborate each component of our method.
	
	\subsection{Notations and Preliminaries}
	Assume that we are given a set of $N$ videos $\mathcal{V} = \left\{\mathcal{V}_i\right\}_{i=1}^N$ and the ground-truth labels $\mathcal{Y} = \left\{\mathcal{Y}_i \right\}_{i=1}^N\in\{1,0\}$. $\mathcal{Y}_i=1$ if an abnormal clip is present in the video and $\mathcal{Y}_i=0$ otherwise. During training, only video-level labels are available. 
	However, in the testing stage, the goal of  this task is to generate frame-level anomaly scores to indicate the temporal location of abnormal events. 
	Following previous MIL-based methods ~\cite{sultani2018real, zhang2019temporal, zhu2019motion, wu2021learning, lv2021localizing}, for each input video $\mathcal{V}_i$, we first divide it into 16-frame non-overlapping $T_i$ clips and use a pre-trained 3D convolutional network to extract features, forming a clip feature sequence $\mathcal{C}_i=\left\{c_{i,1}, c_{i,2}, \ldots, c_{i,T_i}\right\} \in \mathbb{R}^{{T_i} \times D_V}$, where $T_i$ is the number of extracted video clip features and $D_V$ is the feature dimension. 
	Since long untrimmed videos may contain different numbers of clips, which is inconvenient for training.
	Therefore, consistent with \cite{sultani2018real, tian2021weakly}, the video clip features are combined into $S$ temporal video segments denoted as $\mathcal{X}_i=\left\{x_{i,1}, x_{i,2}, \ldots, x_{i,S}\right\} \in \mathbb{R}^{S \times D_V} $ by averaging multiple consecutive clip features. Formally,
	the $s_{th}$ segment feature $x_{i,s}$ is computed as: $x_{i,s}=\frac{1}{c_e-c_s} \sum_{t=c_s}^{c_e} c_{i,t}$,  where $c_s$ and $c_e$ represent the starting index and ending index of the clips contained in the current segment, respectively.
	We treat an abnormal video $\mathcal{V}_i^a$ as a positive bag  and a normal video $\mathcal{V}_i^n$ as a negative bag, and treat each segment $x_{i,s}^a$ or $x_{i,s}^n$ as an instance in the bag. 
	
\subsection{Completeness of Pseudo Labels}\label{sec:com}	
	Inspired by \cite{liu2019completeness}, which uses a diversity loss to model action completeness, we design a completeness enhanced pseudo label generator composed of parallel multi-head classifier, together with a diversity loss to detect as many abnormal events as possible in a video.
	Each head $f_g(\cdot;\phi^k)$ is composed of three fully connected layers parameterized by $\phi^k$. Taking the video segment features $\mathcal{X}_i=\left\{x_{i,s}\right\}_{s=1}^S$ as input, each head outputs the anomaly scores of each segment, which are further passed through a softmax
	to generate a score distribution:
	
	\begin{equation}
	\hat{\mathcal{Y}}_i^k = softmax\left(f_g\left(\mathcal{X}_i; \phi^k\right)\right)
	\end{equation}
	where $\hat{\mathcal{Y}}_i^k \in \mathbb{R}^{S \times 1}$ denotes the score distribution of the $i$-{th} video from the $k$-th head. The predicted score distributions of $K$ heads are then enforced to be distinct from each other by a diversity loss, which minimizes the cosine similarity of the distribution between any two heads:
	\begin{equation}
	\mathcal{L}_{diver}=\frac{1}{Z}  \sum_{k=1}^{K-1} \sum_{q=k+1}^K \frac{{{\hat{\mathcal{Y}}_i^k}} \cdot{{\hat{\mathcal{Y}}_i^q}}}{\left\|{{\hat{\mathcal{Y}}_i^k}}\right\|\left\|{{\hat{\mathcal{Y}}_i^q}}\right\|}
	\end{equation}
	where $Z=\frac{1}{2} K(K-1)$. 
	A regularization term on the norm of the segment score sequences is used to balance multiple heads and to avoid performance degradation due to dominance by one head:
	\begin{equation}
	\mathcal{L}_{norm}=\frac{1}{K} \sum_{k=1}^K \left|\left\|\mathcal{A}^k\right\|-\left\|\mathcal{A}^{avg}\right\|\right|
	\end{equation}
	where $\mathcal{A}^k = f_g\left(\mathcal{X}_i; \phi^k\right)$ denotes the anomaly scores generated by the $k_{th}$ head and $\mathcal{A}^{avg}$ is the average of the anomaly scores produced by each head: $\mathcal{A}^{avg} =\frac{1}{K} \sum_{k=1}^K \left(f_g\left(\mathcal{X}_i; \phi^k\right)\right) $.
	
	Under the action of the diversity loss and norm regularization, the anomaly scores generated by multiple heads can achieve maximum differentiation and detect different anomalous segments. Finally, $\mathcal{A}^{avg}$ is followed by a sigmoid function to obtain the predicted segment-level labels ranging from $0$ to $1$:
	\begin{equation}
	\hat{\mathcal{Y}_i} = sigmoid\left(\mathcal{A}^{avg}\right).
	\end{equation}
	where $\hat{\mathcal{Y}_i}=\left\{\hat{y}_{i,1}, \hat{y}_{i,2}, \ldots, \hat{y}_{i,S}\right\}$ represents the predicted segment-level labels of the $i$-{th} video. For the abnormal video $\mathcal{V}_i^a$, the predicted labels are denoted as $\hat{\mathcal{Y}}_i^a=\left\{\hat{y}_{i,1}^a, \hat{y}_{i,2}^a, \ldots, \hat{y}_{i,S}^a\right\}$. For the normal video $\mathcal{V}_i^n$, the predicted labels are denoted as  $\hat{\mathcal{Y}}_i^n=\left\{\hat{y}_{i,1}^n, \hat{y}_{i,2}^n, \ldots, \hat{y}_{i,S}^n\right\}$.  
	
	Like~\cite{sultani2018real}, the ranking loss  is used to constrain the highest anomaly score of abnormal videos to be higher than that of normal videos:
	\begin{equation}
	\max _{\hat{y}_{i,s}^a \in \hat{\mathcal{Y}}_i^a} \hat{y}_{i,s}^a >\max _{\hat{y}_{i,s}^n \in \hat{\mathcal{Y}}_i^n} \hat{y}_{i,s}^n
	\end{equation}
	To maximize the separability between positive and negative instances, a hinge-based ranking loss is used:
	\begin{equation}
	\mathcal{L}_{MIL}=\max \left(0,1-\max _{\hat{y}_{i,s}^a \in \hat{\mathcal{Y}}_i^a} \hat{y}_{i,s}^a +\max _{\hat{y}_{i,s}^n \in \hat{\mathcal{Y}}_i^n} \hat{y}_{i,s}^n \right)
	\end{equation}

    Finally, the completeness enhanced pseudo label generator is trained with loss as follows:
    \begin{equation}
	\label{Eqn:loss}
	\mathcal{L}_{f_g} = \mathcal{L}_{MIL} + \alpha \mathcal{L}_{diver} + \alpha \mathcal{L}_{norm}
	\end{equation}
	where $\alpha$ is the hyper-parameter to balance the losses. 
    
\subsection{Uncertainty of Pseudo Labels}\label{sec:uncer}	
	Instead of directly using the clip-level pseudo labels obtained in the first stage (Sec.~\ref{sec:com}) to train the final desired clip classifier $f_c$,
	we propose an uncertainty aware self-training strategy to mine clips with reliable pseudo labels. 
	Specifically, we introduce the uncertainty estimation leveraging Monte Carlo (MC) Dropout \cite{gal2016dropout}  
	so that clips with low uncertainty (\emph{i.e.}, reliable) pseudo labels are selected for training $f_c$. 
    This process is conducted several iterations.
	Throughout these iterations, the pseudo labels are continuously refined, eventually generating high-quality fine-grained pseudo labels to train the final desired clip classifier $f_c$. Note that the pseudo labels are initially obtained in the first stage and are then updated by  $f_c$.
	
\noindent \textbf{Uncertainty Estimation.} We use MC-Dropout \cite{gal2016dropout} to estimate the uncertainty of clip-level pseudo labels. For training video clips $\mathcal{C}_i=\left\{c_{i,t}\right\}_{t=1}^{T_i}$, we perform $M$ stochastic forward passes through the model $f$ trained with dropout. In the first iteration, we use the multi-head classifier in the first stage as the trained model (\emph{i.e.}, $f=f_g$). In the remaining iterations, $f=f_c$. Each pass generates clip-level pseudo labels as follows:
	
	\begin{equation}
	\hat{\mathcal{Y}}_i^m=sigmoid\left(f(\mathcal{C}_i; \widetilde{W}^m)\right)
	\end{equation}
	where $\widetilde{W}^m$ denotes the $m^{th}$ sampled masked model parameters and $\hat{\mathcal{Y}}_i^m=\left\{\hat{y}_{i,t}^m\right\}_{t=1}^{T_i}$ . 
    The clip-level pseudo labels 
   $\tilde{\mathcal{Y}}_i=\left\{\tilde{y}_{i,t}\right\}_{t=1}^{T_i}$ 
   used as the supervision for training clip classifier are given by the predictive mean:
    \begin{equation}
    \label{Eqn:E(Y)}
	\operatorname{E}(\hat{\mathcal{Y}}_i) = \frac{1}{M} \sum_{m=1}^M (\hat{\mathcal{Y}}_i^m)
	\end{equation}
    The prediction uncertainties $\mathcal{U}_i=\left\{u_{i,t}\right\}_{t=1}^{T_i}$  of $\hat{\mathcal{Y}}_i$ are given by the predictive variance $\operatorname{Var}(\hat{\mathcal{Y}}_i)$:
	\begin{equation}
    \label{Eqn:Var(Y)}
	\operatorname{Var}(\hat{\mathcal{Y}}_i) = \frac{1}{M} \sum_{m=1}^M \hat{\mathcal{Y}}_i^{m^\top} \hat{\mathcal{Y}}_i^m-E(\hat{\mathcal{Y}}_i)^\top E(\hat{\mathcal{Y}}_i)
	\end{equation}
	
\noindent \textbf{Iterative Reliable Clip Mining.} As the goal is to train a  reliable model with low-uncertainty pseudo labels, we mine reliable clips after uncertainty estimation.
For each video, we rank the uncertainty $\mathcal{U}_i$ of its pseudo labels from small to large and remain the clips with the least $R\%$  uncertainty scores, where $R\%$ represents the sample ratio. 
In this way, reliable video clips and corresponding clip-level pseudo labels can be mined.
	
Since participating in training with only clip-level features ignores the contextual relationship of the video, we use a long-term feature memory \cite{wu2019long} to model the temporal relationship between video clips.
The clip features of each video are stored in a memory pool. After obtaining the selected reliable clip, we retrieve the window size $w$ clip features $\mathcal{H}_{i,t}=[c_{i,t-w},\ldots,c_{i,t-1}]$ before the current clip $c_{i,t}$ from the memory pool. We obtain $\tilde{\mathcal{H}}_{i,t}$ by performing mean pooling on $\mathcal{H}_{i,t}$, and then concatenate the current clip feature $c_{i,t}$ with $\tilde{\mathcal{H}}_{i,t}$ into a new temporal clip feature $\overline{c}_{i,t}$. Thus we can use all reliable temporal features set $\Omega_R(\mathcal{C})$ and reliable clip-level pseudo labels $\Omega_R(\tilde{\mathcal{Y}})$ to train the clip classifier $f_{c}$ based on the binary cross entropy loss:
  
   \begin{equation}
    \begin{split}
    \label{Eqn:CRE}
    \mathcal{L}_{XE}=
    \sum_{\overline{c}_{i,t} \in \Omega_R(\mathcal{C})}-\left({\tilde{y}_{i,t} \log \left(f_c(\overline{c}_{i,t})\right)} \right. \\ 
    \left.{+ (1-\tilde{y}_{i,t}) \log \left(1-f_c(\overline{c}_{i,t})\right)} \right)
    \end{split}
    \end{equation}

\noindent where $\tilde{y}_{i,t} \in \Omega_R(\tilde{\mathcal{Y}}) $
represents the pseudo label of the $t_{th}$ clip in the $i_{th}$ video. Then we can obtain the clip classifier trained in the current iteration, and perform uncertainty estimation and reliable sample selection in the next iteration to further train the desired clip classifier.

    	\begin{algorithm}[t] 
		\caption{Completeness-and-Uncertainty Aware \\ Pseudo Label Enhancement} 
		\label{alg1} 
		\begin{algorithmic}[1] 
			\Require A set of $N$ videos $\mathcal{V}=\left\{\mathcal{V}_i\right\}_{i=1}^N$ and video-level labels $\mathcal{Y}=\left\{\mathcal{Y}_i \right\}_{i=1}^N$.
			
			\Ensure Clip classifier $f_c$.
			
			\State // Completeness Enhanced Pseudo Label Generator.
			\State Extract $T_i$ clip features for each video  $\mathcal{V}_i$ as $\left\{c_{i,t}\right\}_{t=1}^{T_i}$.
			\State Combine $\left\{c_{i,t}\right\}_{t=1}^{T_i}$ into $S$ segment features $\left\{x_{i,s}\right\}_{s=1}^S$.
			\State Training multi-head classifier $f_g$ with $\left\{x_{i,s}\right\}_{s=1}^S$ and $\mathcal{Y}_i$ via Equation \ref{Eqn:loss}.
			\State Obtain the trained model $f=f_g$.
			
			\State // Uncertainty Aware Pseudo Label Refinement.

			\While{ not converged }
            \State Leverage MC-Dropout for the trained model $f$

			\State Compute prediction mean with Equation \ref{Eqn:E(Y)} as the clip-level pseudo labels  
			$\tilde{\mathcal{Y}}_i=\left\{\tilde{y}_{i,t}\right\}_{t=1}^{T_i}$
			
			\State Compute  
			pseudo label uncertainty $\{\mathcal{U}_i\}_{i=1}^N$ via Equation \ref{Eqn:Var(Y)}

			\State $(\mathcal{U}_i)_{sorted}\gets \operatorname{SORT}(\mathcal{U}_i)$. // ascending sort 
			\State $\mathcal{U}^{\prime}_i \gets$ Clip index set of top $R$ percent of $(\mathcal{U}_i)_{sorted}$
	
			\For{$u\gets0$ to $\left|\mathcal{U}^{\prime}_i\right|$}
			\State Obtain reliable clip index $t = \mathcal{U}^{\prime}_i(u)$
            \State Obtain new temporal clip feature $\overline{c}_{i,t}$ 
			\State Add $\overline{c}_{i,t}$ to reliable video clip set   $\Omega_R(\mathcal{C}_i)$
	
		    \State Add $\tilde{y}_{i,t}$ to reliable pseudo label set $\Omega_R(\tilde{\mathcal{Y}}_i)$
			\EndFor
			
			\State All reliable clip set $\Omega_R(\mathcal{C})=\Omega_R(\left\{\mathcal{C}_i\right\}_{i=1}^N)$
		
			\State All reliable label set $\Omega_R(\tilde{\mathcal{Y}})=\Omega_R(\left\{\tilde{\mathcal{Y}}_i\right\}_{i=1}^N)$
		
			\State Training clip classifier $f_c$ with $\Omega_R(\mathcal{C})$ and $\Omega_R(\tilde{\mathcal{Y}})$ via Equation \ref{Eqn:CRE}
			\State Obtain the trained model $f=f_c$
			
			\EndWhile
			\State Return clip classifier $f_c$ for inference.
		\end{algorithmic}
	\end{algorithm}
	
	\subsection{Model Training and Inference} 
	
\noindent	\textbf{Training.} We increase completeness of pseudo labels with Equation \ref{Eqn:loss} to cover as many abnormal clips as possible in the first stage. 
In the second stage, we mine reliable video clips with uncertain estimation to train the clip classifier using Equation \ref{Eqn:CRE} and gradually refine clip-level pseudo labels through multiple iterations. 
Algorithm \ref{alg1} shows the main steps of the training process. 
	
\noindent \textbf{Inference.} Given a test video, we directly utilize the clip classifier $f_{c}$ to predict anomaly scores. 
	
\section{Experimental Results}
We perform experiments on three publicly available datasets including UCF-Crime \cite{sultani2018real}, TAD \cite{lv2021localizing} and XD-Violence \cite{wu2020not}.
	\subsection{Datasets and Evaluation Metrics}
\noindent \textbf{Datasets.}	UCF-Crime is a large-scale benchmark dataset for video anomaly detection with 13 anomaly categories. The videos are captured from diverse scenes, such as streets, family rooms, and shopping malls. The dataset contains 1610 training videos annotated with video-level labels and 290 test videos with frame-level annotation. 
TAD is a recently released dataset in traffic scenario. It covers 7 real-world anomalies in 400 training videos and 100 test videos. In line with the weak supervision setting in \cite{sultani2018real}, the training set is annotated with video-level labels, and test set provides frame-level labels.
XD-Violence is the largest dataset currently used for weakly supervised anomaly detection. Its videos are collected through multiple channels, such as movies, games, and car cameras. It contains 3954 training videos with video-level labels and 800 test videos with frame-level labels, covering 6 anomaly categories. Moreover, it provides audio-visual signals, enabling anomaly detection by leveraging multimodal cues.
	
\noindent \textbf{Evaluation Metrics.} Similar to the previous works \cite{sultani2018real, zhang2019temporal, zhong2019graph, feng2021mist}, we choose the area under the curve (AUC) of the frame-level receiver operating characteristic curve to evaluate the performance of our proposed method on UCF-Crime and TAD datasets. For XD-Violence, following \cite{tian2021weakly, wu2020not, wu2021learning, li2022self}, we use average precision (AP) as the metric.
	
\subsection{Implementation Details}
\noindent \textbf{Feature Extractor.} Consistent with existing methods \cite{tian2021weakly},  we use pre-trained I3D \cite{carreira2017quo} model to extract clip features from 16 consecutive frames. Then we divide each video into 32 segments as the input to the multi-head classifier, \emph{i.e.}, $S=32$.  For the XD-violence dataset, following the setting in \cite{wu2020not, wu2021learning}, we extracted audio features by leveraging the VGGish network \cite{gemmeke2017audio}.
	
\noindent \textbf{Training Details.} The head number $K$ of the multi-head classifier is 2 for all datasets. Each head consists of three fully connected (FC) layers with 512, 128, and 1 units, respectively. The first and third fully connected layers are followed by ReLU activation and sigmoid activation, respectively. We use a dropout function between FC  layers with a dropout rate of $0.6$. The classifier is trained using the Adadelta optimizer with a learning rate of 0.1. The parameter $\alpha$ is set to $10$ for UCF-Crime, and 0.1 for TAD and XD-Violence. The number of stochastic forward passes $M$ is set to $50$ for all datasets when generating the initial pseudo labels and uncertainty scores. During pseudo label refinement, the clip classifier consists of three FC layers and 80\% of the dropout regulation is used between FC layers. It is trained using the Adam optimizer with a learning rate of $1e-4$ and a weight decay of $5e-4$. At the end of each iteration, we set $M=5$ to obtain pseudo labels and uncertainty scores.

	\subsection{Comparisons with Prior Work}
	
\begin{table}[t]
\centering
\begin{tabular}{cccc}
			\toprule
			Supervised   &    Methods &   Feature & AUC(\%) \\
			\midrule
			\multirow{4}* {Semi} & Hasan \emph{et al.} \cite{hasan2016learning} &  AE & $50.60$ \\
			~ & Ionescu \emph{et al.} \cite{ionescu2019object} &   - & $61.60$ \\
			~ & Lu \emph{et al.} \cite{lu2013abnormal} &   Dictionary & $65.51$ \\
			~ & Sun \emph{et al.} \cite{sun2020scene} &    - & $72.70$ \\
			\midrule
			\multirow{15}* {Weakly} & Binary classiﬁer  &  C3D & $50.00$ \\
			~ & Sultani \emph{et al.} \cite{sultani2018real} &  C3D & $75.41$ \\
			~ & Zhang \emph{et al.} \cite{zhang2019temporal}&  C3D &  $78.66$ \\
			~ & Zhu \emph{et al.} \cite{zhu2019motion}&   AE  & $79.00$ \\
			~ & GCN \cite{zhong2019graph}&   TSN  & $82.12$ \\
			~ & CLAWS \cite{zaheer2020claws}&   C3D & $83.03$ \\
            ~ & Wu \emph{et al.} \cite{wu2020not}& I3D & $82.44$ \\
			~ & MIST \cite{feng2021mist}&  I3D & $82.30$ \\
            ~ & Wu \emph{et al.} \cite{wu2021learning}& I3D & $84.89$ \\
			~ & WSAL \cite{lv2021localizing}&  TSN & $85.38$ \\
			~ & RTFM \cite{tian2021weakly}&  I3D & $84.30$ \\
			~ & MSL \cite{li2022self}&  I3D  & $85.30$ \\
            ~ & BN-SVP \cite{sapkota2022bayesian}&  I3D  & $83.39$ \\
			~ & Ours &  I3D   & \textbf{86.22} \\
			\bottomrule
		\end{tabular}
		\caption{Comparison with other methods on UCF-Crime dataset.}
		\label{tab:UCF_Res}
	\end{table}
	
\noindent \textbf{Results on UCF-Crime.} Table \ref{tab:UCF_Res} summarizes the performance comparison between the proposed method and other methods on UCF-Crime test set. From the results, we can see that our proposed method outperforms all  the previous semi-supervised methods \cite{hasan2016learning, ionescu2019object, lu2013abnormal, sun2020scene} and weakly supervised methods \cite{sultani2018real, zhang2019temporal, zhu2019motion, zhong2019graph, zaheer2020claws, wu2020not, feng2021mist, wu2021learning, lv2021localizing, tian2021weakly, li2022self, sapkota2022bayesian}. Using the same I3D-RGB features, an absolute gain of $0.92\%$ is achieved in terms of the AUC when compared to the best previous two-stage method \cite{li2022self}.
	
\noindent \textbf{Results on TAD.} The performance comparisons on TAD dataset are shown in Table \ref{tab:TAD_Res}. Compared with the previous semi-supervised approaches \cite{luo2017revisit, liu2018future} and weakly supervised methods \cite{sultani2018real, zhu2019motion, feng2021mist, lv2021localizing, tian2021weakly}, our method achieves the superior performance on AUC. Remarkably, with the same I3D features,  our method is $2.40\%$ and $2.02\%$ better than two-stage MIST \cite{feng2021mist} and one-stage RTFM \cite{tian2021weakly}.

\begin{table}[t]
	\centering
	\begin{tabular}{cccc}
		\toprule
		Supervised   &    Methods &   Feature & AUC(\%) \\
		\midrule
		\multirow{2}* {Semi} & Luo \emph{et al.} \cite{luo2017revisit} &   TSN & $57.89$ \\
		~ & Liu \emph{et al.} \cite{liu2018future} &   - & $69.13$ \\
		\midrule
		\multirow{8}* {Weakly} & Sultani \emph{et al.} \cite{sultani2018real} &  C3D & $83.27$ \\
		~ & Sultani \emph{et al.} \cite{sultani2018real} &  TSN & $85.95$ \\
		~ & Sultani \emph{et al.} \cite{sultani2018real} &  I3D & $88.34$ \\
		~ & Zhu \emph{et al.} \cite{zhu2019motion}&   TSN & $83.08$ \\
		~ & MIST \cite{feng2021mist}&  I3D & $89.26$ \\
		~ & WSAL \cite{lv2021localizing}&   TSN & $89.64$ \\
		~ & RTFM \cite{tian2021weakly}&  I3D & $89.64$ \\
		~ & Ours &  I3D & \textbf{91.66} \\
		\bottomrule
	\end{tabular}
	\caption{Comparison with other methods on TAD dataset.}
	\label{tab:TAD_Res}
\end{table}

\begin{table}[t]
	\centering
	\begin{tabular}{cccc}
		\toprule
		Supervised   &    Methods &   Feature & AP(\%) \\
		\midrule
		\multirow{3}* {Semi} & SVM baseline & I3D+VGGish & $50.78$ \\
		~ & OCSVM \cite{scholkopf1999support} & I3D+VGGish & $27.25$ \\
		~ & Hasan \emph{et al.} \cite{hasan2016learning} & I3D+VGGish & $30.77$ \\
		\midrule
		\multirow{8}* {Weakly} & Sultani \emph{et al.} \cite{sultani2018real} &  I3D+VGGish & $73.20$ \\
		~ & Wu \emph{et al.} \cite{wu2020not}& I3D & $75.41$ \\
		~ & Wu \emph{et al.} \cite{wu2020not}& I3D+VGGish & $78.64$ \\
		~ & Wu \emph{et al.} \cite{wu2021learning}& I3D+VGGish & $75.90$ \\
		~ & RTFM \cite{tian2021weakly}&  I3D  & $77.81$ \\
		~ & MSL \cite{li2022self}&  I3D  & $78.28$ \\
		~ & Ours &  I3D  & $78.74$ \\
		~ & Ours &  I3D+VGGish  & \textbf{81.43} \\
		\bottomrule
	\end{tabular}
	\caption{Comparison with other methods on XD-Violence dataset.}
	\label{tab:Violence_Res}
\end{table}

\begin{table}[t]
	\centering
	\setlength{\tabcolsep}{2mm}{
		\begin{tabular}{cccc}
			\toprule
			Baseline   &  Completeness  &   Uncertainty & AUC(\%) \\
			\midrule
			$\surd$ & ~ & ~ & $82.86$ \\
			$\surd$ & $\surd$ & ~ & $84.89$ \\
			$\surd$ & ~ & $\surd$ & $84.69$ \\
			$\surd$ & $\surd$ & $\surd$ & \textbf{86.22} \\
			\bottomrule
	\end{tabular}}
	\caption{Ablation results on UCF-Crime dataset.}
	\label{tab:Ablation_Res}
\end{table}
	
\noindent \textbf{Results on XD-Violence.} As shown in Table \ref{tab:Violence_Res}, our method also achieves favorable performance when compared with previous semi-supervised approaches \cite{scholkopf1999support, hasan2016learning} and weakly supervised methods \cite{sultani2018real, wu2020not, wu2021learning, tian2021weakly, li2022self} on XD-Violence dataset. Using the same I3D features, our method achieves new state-of-the-art performance of $78.74\%$ AP. Following Wu \emph{et al.} \cite{wu2020not}, we show the results when using the fused features of I3D and VGGish. Our method gains clear improvements against Wu \emph{et al.} \cite{wu2020not} by $2.79\%$ in terms of AP. Moreover, the comparison results between using only I3D features and using multi-modal features (\emph{i.e.}, I3D and VGGish) demonstrate that audio can provide more useful information for anomaly detection.

	\subsection{Ablation Study}
	We conduct multiple ablation studies on UCF-Crime dataset to analyze how each component in our method influences the overall performance and show the results in Table \ref{tab:Ablation_Res}. We start with the baseline that directly uses the MIL-based method as a pseudo label generator and then selects all the pseudo labels to train a clip-level classifier.
	
\noindent \textbf{Effectiveness of Completeness of Pseudo Labels.} To demonstrate the necessity of exploiting the completeness property, we compare our method that using multi-head classifier constrained by diversity loss as pseudo label generator (the $2^{nd})$ row) with baseline (the $1^{st}$ row) on UCF-Crime dataset. The results show that the completeness property of pseudo labels can achieve $2.03\%$ improvement in terms of AUC, which proves that taking the completeness of pseudo labels into consideration can effectively enhance the quality of pseudo labels and thus improve the anomaly detection performance.
	
\noindent \textbf{Effectiveness of Uncertainty of Pseudo Labels.} To investigate the effect of exploiting the uncertainty property, we conduct experiments (the $3^{rd}$ row) with only iterative uncertainty aware pseudo label refinement strategy added to the baseline. From the results, we can see that uncertainty property can bring a $1.83\%$ performance gain in AUC, which indicates that using an uncertainty aware self-training strategy to gradually improve the quality of pseudo labels is important.

\begin{figure}[!t]
	\centering
	\includegraphics[width=\linewidth,trim=170 85 180 80,clip]{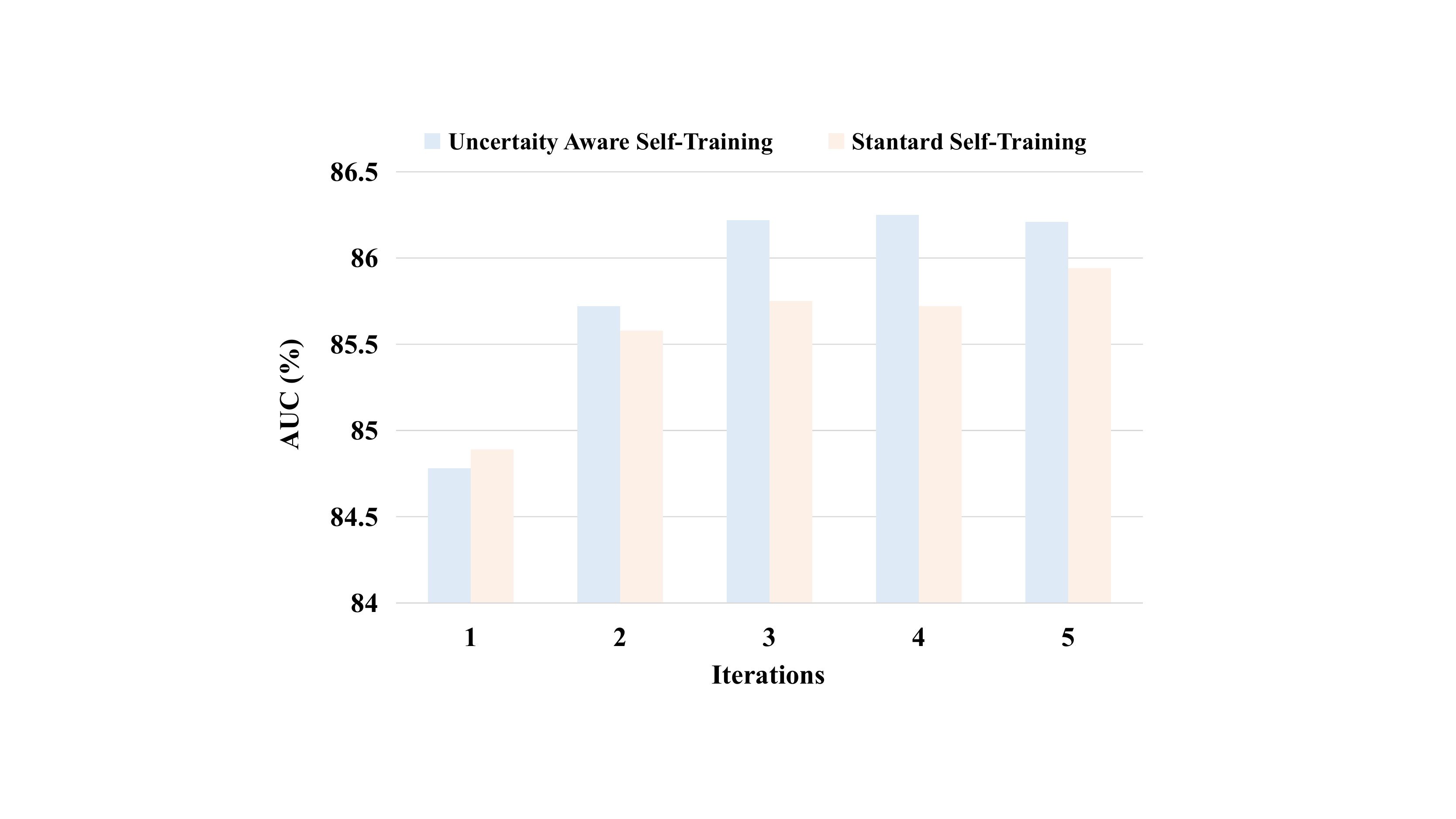}
	\caption{Performance comparison of training strategy between Uncertainty Aware Self-Training and Standard Self-Training.}
	\label{fig:self-training}
\end{figure}
	
\noindent \textbf{Analysis on the Uncertainty Aware Self-Training.} To look deeper into the proposed uncertainty aware self-training strategy, we also make a comparison with standard self-training mechanism that randomly selects samples from the new labeled set. 
As the number of iterations increases, the performance of the uncertainty aware self-training strategy increases rapidly and then stabilizes gradually, whereas the performance improvement of the standard self-training strategy is much smaller.
This shows that adopting the uncertainty-aware self-training strategy, that is, using uncertainty estimation to select reliable samples for training the desired classifier in each iteration is more beneficial to the quality of pseudo labels.

    \begin{table}[!t]
		\centering
		\begin{tabular}{cccccc}
			\toprule
			$K$   &  $1$  &   $2$  &  $3$ & $4$ & $5$ \\
			\midrule
			AUC(\%) & $84.69$ & \textbf{86.22} & $85.05$ & $85.90$ & $85.10$ \\
			\bottomrule
		\end{tabular}
		\caption{Performance comparison of different head numbers.}
		\label{tab:head_num}
	\end{table}

    \begin{table}[!t]
		\centering
		\begin{tabular}{cccccc}
			\toprule
			$\alpha$   &  $0.01$  &   $0.1$  &  $1$ & $10$ & $100$ \\
			\midrule
			AUC(\%) & $84.56$ & $84.43$ & $84.82$ & \textbf{86.22} & $75.06$ \\
			\bottomrule
		\end{tabular}
		\caption{Performance comparison of different diversity loss weights.}
		\label{tab:Loss_weight}
	\end{table}

	\subsection{Hyperparameter Analysis}
 
\noindent \textbf{Effect of the head number $K$.} In Table \ref{tab:head_num}, we show the comparative result of varying the head number $K$ of the multi-head classifier on UCF-Crime dataset. When $K=1$, the pseudo label generator is a single-head classifier, which limits the ability of pseudo labels to cover complete anomalies. As the head number increase from $2$ to $5$, the performance of multi-head classifier models all exceed that of single-head classifier model and our method yields the best performance when $K=2$. Since the performance difference between models with head numbers from $2$ to $8$ is not so significant, we set $K=2$ for all datasets.

\noindent \textbf{Effect of the diversity loss weight $\alpha$.} To explore the influence of the diversity loss weight, we conduct experiments on UCF-Crime dataset and report the AUC with different diversity loss weights. As shown in Table \ref{tab:Loss_weight}, we report the results as $\alpha$ increases from $0.01$ to $100$. The AUC can be consistently improved as $\alpha$ grows from $0.01$ to 10 and then decreases when $\alpha$ increases to $100$,  which means $10$ is sufficient to model the completeness of pseudo labels on UCF-Crime dataset. For the TAD and XD-Violence datasets, the weight of the diversity loss $\alpha$ is set to $0.1$.

\noindent \textbf{Effect of self-training iterations $j$.} In Figure \ref{fig:self-training}, we report the AUC of our proposed method with different iteration on UCF-Crime dataset. In general,  we observe the performance to improve rapidly initially as $j$ increases , and gradually converge in $3-5$ iterations. Considering efficiency and performance comprehensively, we set $j=3$ for UCF-Crime and XD-Violence datasets. Due to the relatively small scale of the TAD dataset, we set $j=5$ for TAD. 
	
\begin{figure}[!t]
		\centering
		\includegraphics[width=0.9 \linewidth, trim=160 60 150 60,clip]{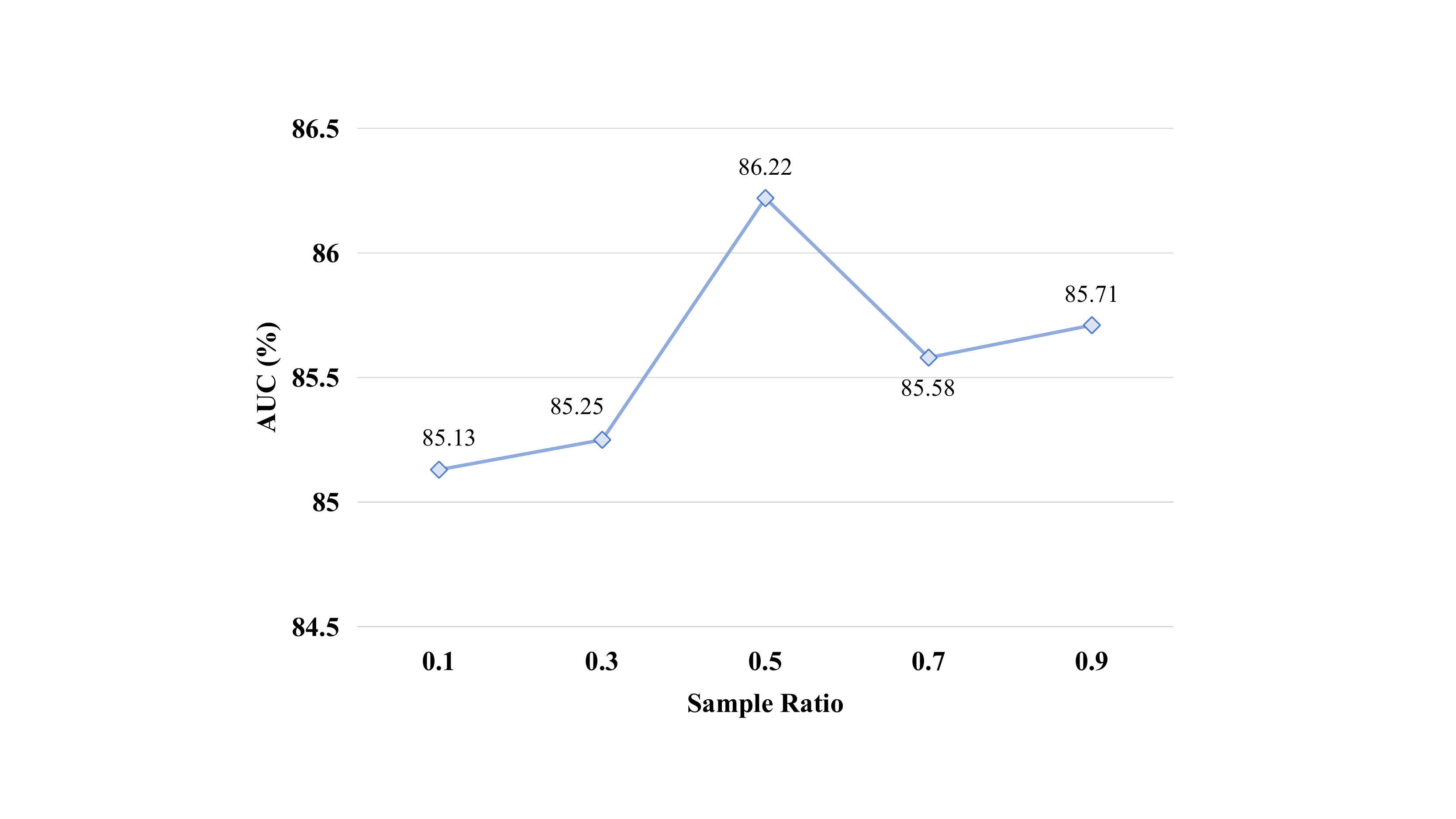}
		\caption{Performance comparison of different sample ratios.}
		\label{fig:sample_ratio}
	\end{figure}

\noindent \textbf{Effect of the sample ration $R$.} Figure \ref{fig:sample_ratio} reports the experimental results evaluated with different sample ratios on UCF-Crime dataset. We observe that setting $R$ to $0.5$ is an optimum choice. As $R$ increases, the AUC improves rapidly and then declines gradually. According to our analysis, the sampling ratio of reliable samples should depend on the scale of the training set. With large-scale training set, it is more likely to generate high-quality pseudo labels, so a larger sampling ratio can be set. On the contrary, a smaller sampling ratio should be set. Therefore, for the large-scale XD-Violence dataset, we set $R=0.7$. For the small-scale TAD dataset, R is set to $0.3$.

	\begin{figure}[!t]
		\centering
		
		\subcaptionbox{Explosion\label{fig:Explosion}}{\includegraphics[width=\mysize]{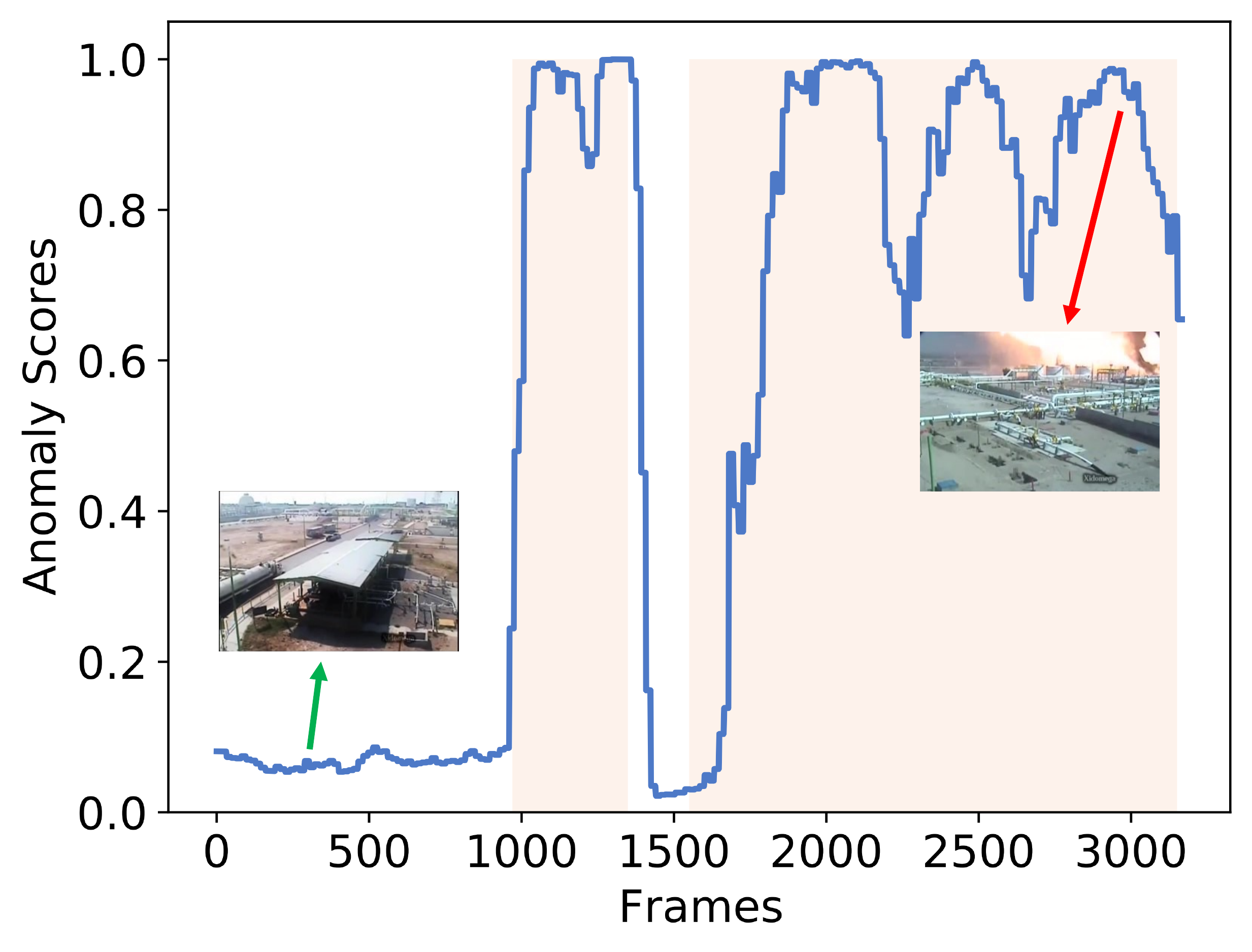}}
		\subcaptionbox{PedestrianOnRoad\label{fig:PedestrianOnRoad}}{\includegraphics[width=\mysize]{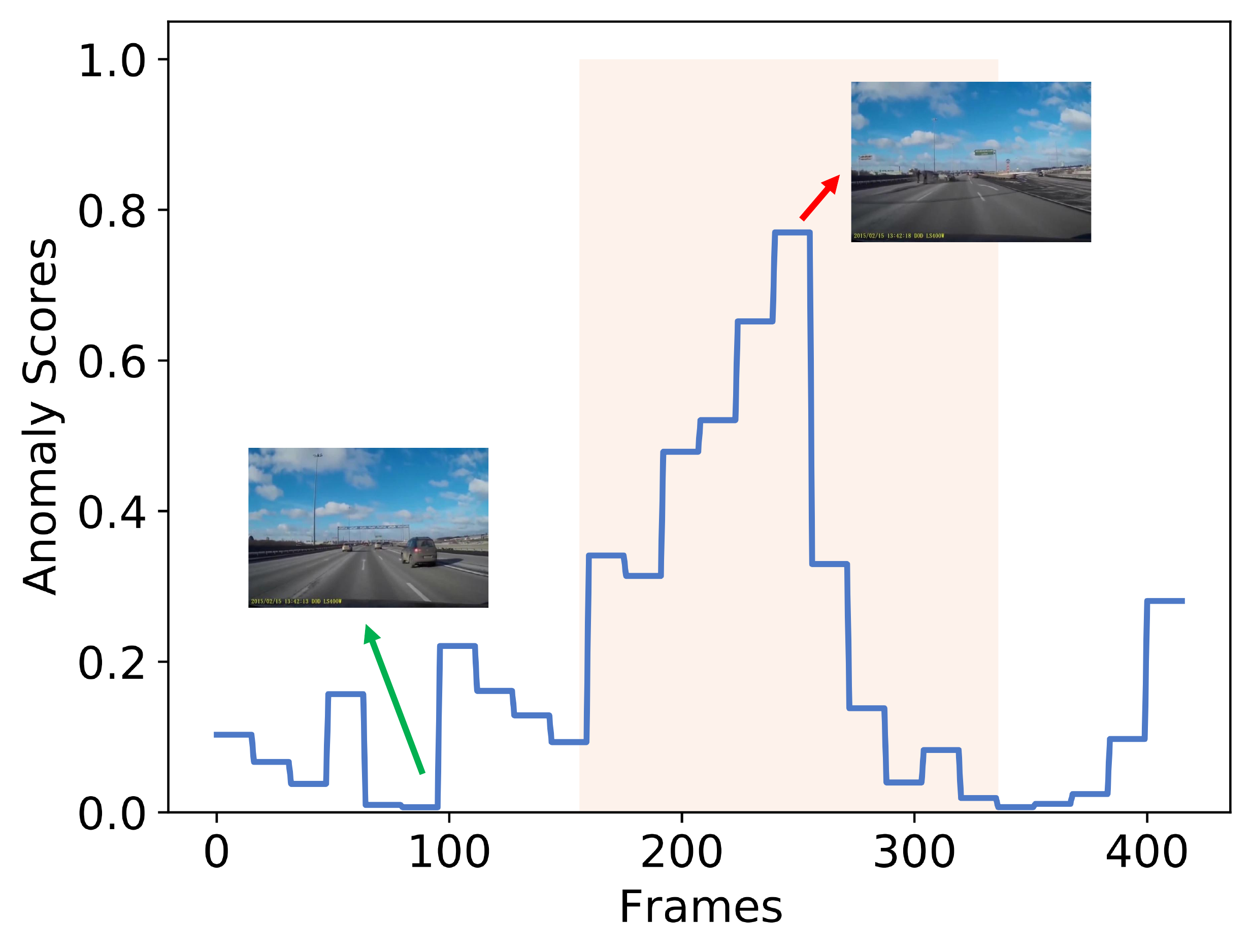}}
		\subcaptionbox{Fighting\label{fig:Fighting}}{\includegraphics[width=\mysize]{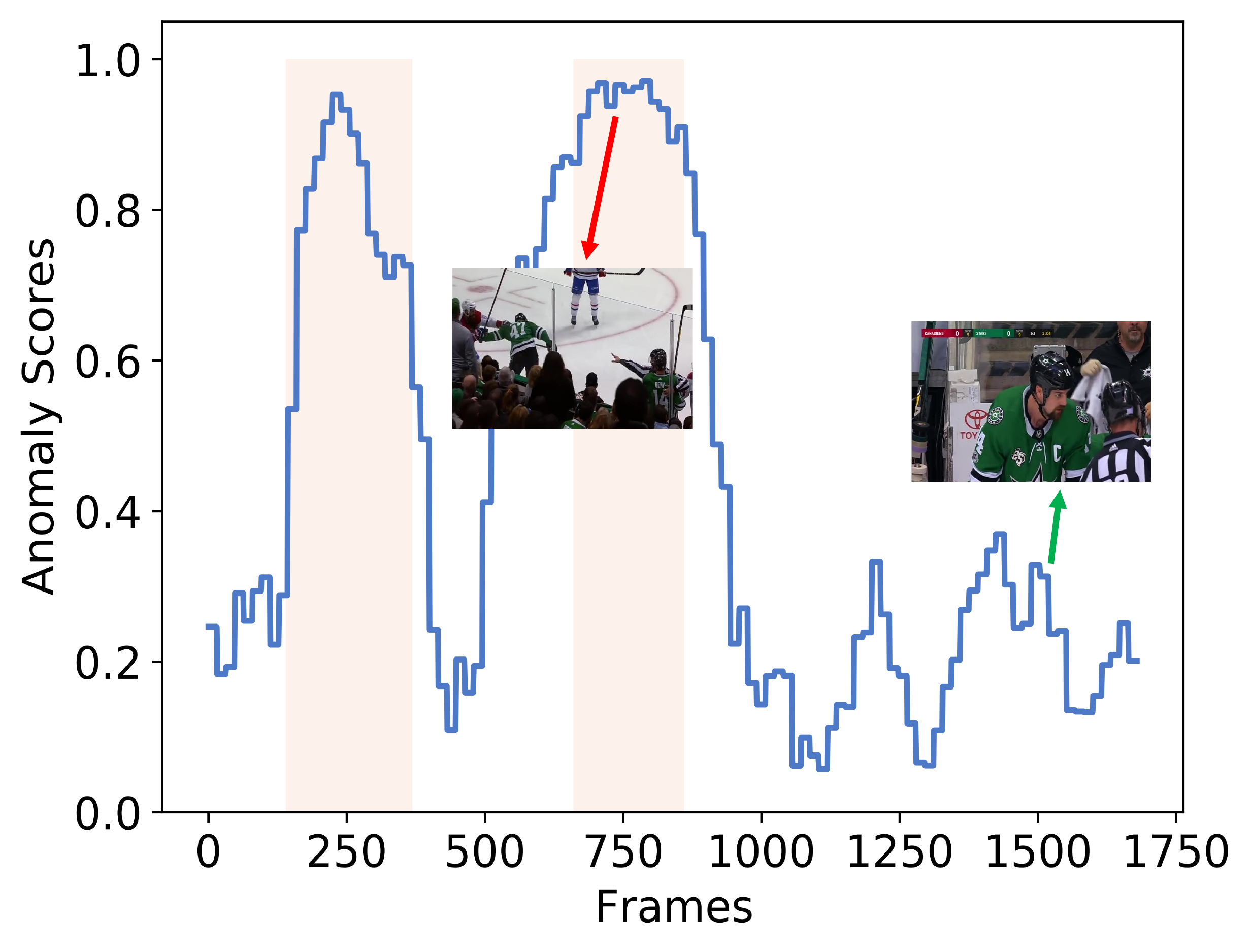}}
		\subcaptionbox{Normal\label{fig:violence_nor}}{\includegraphics[width=\mysize]{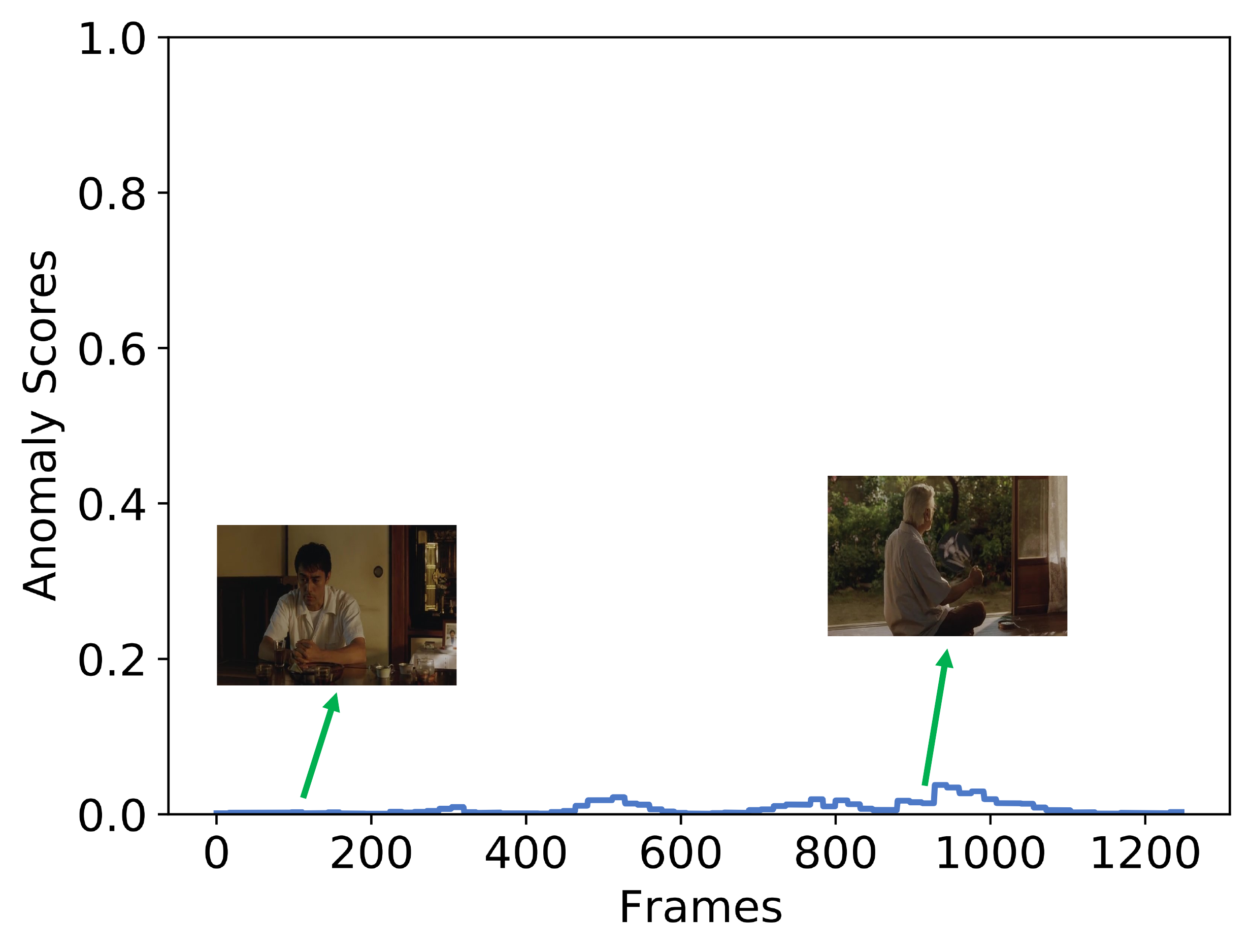}}
		
		\caption{Qualitative results on UCF-Crime (Explosion), TAD (PedestrianOnRoad) and XD-Violence (Fighting, Normal) test videos. The pink square area is the interval where abnormal events occur in the video, blue curve indicates the predicted anomaly scores of video frame.}
		\label{fig:anomaly_score}
	\end{figure}
	
	\subsection{Qualitative Results}
	In order to further prove the effectiveness of our method, we visualize the anomaly score results on UCF-Crime, TAD and XD-Violence datasets in Figure \ref{fig:anomaly_score}. As shown in Figure \ref{fig:Explosion} and Figure \ref{fig:PedestrianOnRoad}, our method can predict relatively accurate anomaly scores for the multi-segment abnormal event (Explosion) in UCF-Crime dataset and the long-term abnormal event (PedestrianOnRoad) in TAD dataset. Figure \ref{fig:Fighting} and Figure \ref{fig:violence_nor} depict the anomaly scores of the abnormal event (Fighting) and normal event in XD-Violence dataset, our method can completely detect two anomalous intervals in the abnormal video and predict anomaly scores close to 0 for the normal video.

	\section{Conclusions}
	In this paper, we focus on enhancing the quality of pseudo labels and propose a two-stage self-training method that exploits completeness and uncertainty properties. First, to enhance the completeness of pseudo labels, a multi-head classification module constrained by a diversity loss is designed to generate pseudo labels that can cover as many anomalous events as possible. Then, an iterative uncertainty-aware self-training strategy is employed to select reliable samples to train the clip classifier. The output of the clip classifier is gradually refined through multiple iterations, resulting in high-quality pseudo labels to train the final desired clip classifier. Experiments on UCF-Crime, TAD and XD-Violence datasets show the 
favorable performance of our method. Extensive ablation studies also demonstrate that it is effective to exploit the completeness and uncertainty properties of pseudo labels for weakly supervised video anomaly detection.

	{\small
		\bibliographystyle{ieee_fullname}
		\bibliography{egbib}
	}
	
\end{document}